\documentclass[journal=jacsat,manuscript=article]{achemso}
\usepackage{float}
\usepackage[version=3]{mhchem} % Formula subscripts using \ce{}
\usepackage[version=3]{mhchem} % Formula subscripts using \ce{}
\usepackage{adjustbox}
\usepackage{xcolor}
\usepackage{array}
\usepackage{multirow}
\usepackage{graphicx}
\usepackage{makecell}
\usepackage{cellspace}
\usepackage{booktabs}
\usepackage{tabularx}
\usepackage{placeins} 
\usepackage{float} 
\usepackage{longtable}
\usepackage{float}
\DeclareUnicodeCharacter{2212}{-}

\author{Ying-Ting Yeh}
\altaffiliation{These authors contributed equally to this work.}
\author{Janghoon Ock}
\altaffiliation{These authors contributed equally to this work.}
\affiliation[ChemE]
{Department of Chemical Engineering, Carnegie Mellon University, 5000 Forbes Avenue, Pittsburgh, PA 15213, USA}
\alsoaffiliation[unl-cheme]
{Department of Chemical and Biomolecular Engineering, University of Nebraska--Lincoln, Lincoln, NE 68588, USA}
\author{Achuth Chandrasekhar}
\affiliation{Department of Mechanical Engineering, Carnegie Mellon University, 5000 Forbes Avenue, Pittsburgh, PA 15213, USA}
\author{Shagun Maheshwari}
\affiliation{Department of Material Science  Engineering, Carnegie Mellon University, 5000 Forbes Avenue, Pittsburgh, PA 15213, USA}
\author{Amir Barati Farimani}
\email{barati@cmu.edu}
\affiliation[MechE]
{Department of Mechanical Engineering, Carnegie Mellon University, 5000 Forbes Avenue, Pittsburgh, PA 15213, USA}

\title[An \textsf{achemso} demo]
  {Text to Band Gap: Pre-trained Language Models as Encoders for Semiconductor Band Gap Prediction}

\abbreviations{IR,NMR,UV}

\begin{document}

% \begin{tocentry}

% \centering
% \includegraphics{Figure/TOC.pdf} 
% \label{fig:toc}

% \end{tocentry}

\begin{abstract}
We investigate transformer-based language models, including RoBERTa, T5, Llama-3, and MatSciBERT, for predicting the band gaps of semiconductor materials directly from textual descriptions. The inputs encode key material features, such as chemical composition, crystal system, space group, and other structural and electronic properties. Unlike shallow machine learning models, which require extensive feature engineering, or Graph Neural Networks, which rely on graph representations derived from atomic coordinates, pretrained language models can process textual inputs directly, eliminating the need for manual feature preprocessing or structure-based encoding. Material descriptions were constructed in two formats: structured strings with a consistent template and natural language narratives generated via the ChatGPT API. Each model was augmented with a custom regression head and finetuned for band gap prediction task. Language models of different architectures and parameter sizes were all able to predict band gaps from human-readable text with strong accuracy, achieving MAEs in the range of 0.25–0.33 eV, highlighting the success of this approach for scientific regression tasks. Finetuned Llama-3, with 1.2 billion parameters, achieved the highest accuracy (MAE 0.248 eV, $R^2$ 0.891). MatSciBERT, pretrained on materials science literature, reached comparable performance (MAE 0.288 eV, $R^2$ 0.871) with significantly fewer parameters (110 million), emphasizing the importance of domain-specific pretraining. Attention analysis shows that both models selectively focus on compositional and spin-related features while de-emphasizing geometric features, reflecting the difficulty of capturing spatial information from text. These results establish that pretrained language models can effectively extract complex feature-property relationships from textual material descriptions, providing a scalable, language-native framework for materials informatics.
\end{abstract}

%Embedding visualizations further reveal that fine-tuning aligns material representations according to crystal system and composition, demonstrating that the models learn meaningful feature-property correlations. 

\section{Introduction}
The band gap of semiconductor materials is a fundamental property that directly impacts their electronic and optical behaviors. This parameter dictates crucial attributes such as conductivity, light absorption, and emission, making it essential for the performance of various electronic, optoelectronic, and photovoltaic devices \cite{yu2010fundamentals}. Therefore, the precise prediction and control of the band gap are vital for optimizing semiconductor applications in these fields\cite{kim2020band}.

Band gaps are determined primarily through experimental methods, with UV-visible absorption spectroscopy and photoluminescence spectroscopy being the most commonly used techniques\cite{masood2023enhancing}. However, these experimental methods can only measure the band gaps of synthesized materials and are not applicable to new materials designed theoretically. On the computational side, Density Functional Theory (DFT) has been the primary tool for studying the electronic structure of materials\cite{koch2015chemist,kohn1965self}. While DFT can accurately simulate electronic properties such as band structures and band gaps, its high computational cost and resource-intensive nature make it less practical for high-throughput material screening, especially for complex systems\cite{schleder2019dft}. 

Machine learning (ML) methods have become powerful tools for addressing the computational challenges of DFT. Shallow ML models, such as Random Forest and Support Vector Regression, are commonly used to predict materials properties like band gaps based on material descriptors \cite{wang2021accurate, rajan2018machine, zhuo2018predicting}. These models offer a cost-effective alternative to solving the full quantum mechanical equations, significantly reducing computational overhead. However, these models often struggle with non-numerical features, requiring extensive preprocessing to convert material properties into numerical formats. This reliance on extensive preprocessing and feature engineering not only adds complexity but also risks discarding nuanced or qualitative information, such as symmetry, bonding environments, or textual metadata, that could be valuable for accurate property prediction~\cite{faber2017prediction, choudhary2021atomistic}. Recent advances in deep learning, particularly Graph Neural Networks (GNNs) \cite{taniguchi2023graph}, have significantly enhanced the modeling of structure–property relationships by representing materials as atomic graphs that encode interatomic connectivity. However, these methods inherently rely on explicit structural information and require the conversion of atomic configurations into graph representations, introducing additional preprocessing steps and limiting their applicability to systems with well-defined crystal structures. Furthermore, GNNs still face limitations in integrating non-numerical properties, such as compound names, into the training process without additional preprocessing steps. These challenges underscore the need for approaches that can seamlessly handle both numerical and non-numerical features in material property predictions while minimizing complex preprocessing requirements.

Language models offer unique advantages by directly utilizing human-readable text data, eliminating the need for elaborate pre-processing while preserving critical information embedded in material descriptions~\cite{ock2023catberta, ock2024multimodal, peptidebert, pak2025}. This streamlines the prediction process. Recent advances in natural language processing, particularly with large language models (LLMs), have introduced transformative possibilities for materials science. This simplifies the prediction process compared to conventional ML approaches, which typically require precise atomic coordinates or extensive preprocessing to generate numerical features. In contrast, LLMs can directly process text-based descriptions. Leveraging this capability, we predict band gap values directly from text-formatted input, bypassing the need for detailed structural data and eliminating complex feature engineering.

Recent advances in natural language processing, particularly with LLMs, have introduced transformative opportunities for materials science. For instance, AlloyBERT demonstrates the potential of transformer-based models to predict material properties from descriptive text~\cite{chaudhari2024alloybert}. Similarly, AMGPT showcases the benefits of using composition-based input strings and finetuned LLMs, enabling accurate and efficient predictions for materials science tasks~\cite{jacobs2024regression, chandrasekhar2024amgpt}. Additionally, CatBERTa, a RoBERTa-based predictive model, has been developed to predict adsorption energy in catalyst systems~\cite{ock2023catberta, ock2024multimodal}.

% contributions
In this study, we explore the use of transformer-based language models, RoBERTa, T5, Llama-3, and MatSciBERT to predict the band gaps of semiconductor materials directly from textual descriptions. These models enable the direct encoding of structured or natural language representations of materials, such as chemical composition, crystal symmetry, and electronic features, without requiring conventional feature engineering. While pretrained language models possess strong linguistic priors, we emphasize that fine-tuning on domain-specific objectives is essential for adapting them to materials property prediction. We implement task-specific regression heads on top of each model and finetune them on a curated dataset of inorganic compounds. This approach allows models to learn mappings from text-based input to scalar band gap values. These models provide a flexible framework for property prediction from textual materials data, extending the application of language models beyond conventional natural language tasks into scientific domains such as materials informatics.

\section{Methods}
\subsection{RoBERTa}

RoBERTa (Robustly Optimized BERT Approach) is an encoder-only transformer that refines BERT’s pretraining strategy through dynamic masking and longer sequence training~\cite{liu2019roberta}. In this work, we employ the RoBERTa model (12 layers, 12 attention heads, 768 hidden units) as one of the backbone architectures for band gap prediction. Its bidirectional attention enables effective contextual encoding of structured text inputs derived from material features. Rather than emphasizing linguistic tasks, our framework fine-tunes RoBERTa to learn correlations between textual feature descriptions and scalar band gap values. This objective-specific adaptation allows the model to repurpose general language representations for scientific property prediction.

\subsection{T5}

T5 (Text-to-Text Transfer Transformer) is a unified encoder–decoder architecture that reformulates all NLP tasks as text-to-text transformations~\cite{raffel2020exploring}. We employ the T5-small (6 encoder and 6 decoder layers, 512 hidden size, 8 attention heads) to explore how a sequence-to-sequence framework can be applied to scientific regression problems. Throughout this manuscript, we refer to this model simply as T5. 

Pretrained with a span corruption objective, T5 learns to reconstruct masked text segments, promoting stronger global reasoning than token-level masking. In our framework, we utilize only the encoder output and append a custom regression head for band gap prediction. This setup leverages T5’s contextual encoding ability while avoiding unnecessary generative complexity, allowing efficient adaptation to structured material descriptions and scalar property prediction.

\subsection{Llama-3}

Llama-3 is a family of decoder-only transformer models developed by Meta, designed to provide state-of-the-art performance with efficient scaling across parameter sizes~\cite{meta_llama_32,meta_llama_31}. In this work, we specifically use the lightweight Llama-3.2-1B variant (approximately one billion parameters, embedding size 2048), which offers a strong balance between performance and computational efficiency. Throughout this manuscript, we refer to this model simply as Llama-3. 

We adapt Llama-3 by attaching a custom regression head to the decoder output for band gap prediction. Architectural components such as RMSNorm, SwiGLU activation, and Rotary Positional Embeddings improve training stability and contextual reasoning, while the SentencePiece-based tokenizer effectively encodes scientific symbols and numerical expressions. This configuration enables Llama-3 to extend its pretrained linguistic representations toward modeling structured material descriptions for quantitative property prediction.

\subsection{MatSciBERT}

MatSciBERT is a BERT-based language model pretrained on materials science literature to capture domain-specific terminology, chemical formulas, and structural descriptions~\cite{gupta2022matscibert}. It uses the standard BERT-base architecture with 12 encoder layers, 768-dimensional embeddings, 12 attention heads, and a 3072-dimensional feed-forward layer, while its masked language modeling pretraining is optimized for the symbolic and long-tail vocabulary common in scientific texts. This provides stronger representations for materials-related language than general-purpose models.

In our framework, we employ the MatSciBERT encoder with a custom regression head to predict band gaps from structured material descriptions. This design leverages MatSciBERT’s domain-aware embeddings while maintaining architectural simplicity, allowing efficient adaptation to property prediction tasks within materials informatics.

\subsection{Shallow ML Models}

We implemented three conventional regression algorithms, Random Forest (RF), Support Vector Regression (SVR), and Extreme Gradient Boosting (XGBoost), using the same dataset of structured material descriptors. All models were trained to predict the band gap values from numerical features extracted from the materials database.

The RF model was implemented using scikit-learn’s \texttt{RandomForestRegressor}. The model constructs an ensemble of decision trees trained on randomly sampled subsets of both data and features, and outputs the average prediction across all trees. A five-fold cross-validated grid search was used to optimize key hyperparameters. The final configuration employed 1,000 trees, a maximum depth of 50, a minimum of two samples per split, and two samples per leaf. 

For the SVR, we used scikit-learn’s \texttt{SVR} implementation with a radial basis function (RBF) kernel. The model projects the input features into a high-dimensional space where a linear regression is performed within an $\epsilon$-insensitive margin. Grid search with five-fold cross-validation determined the optimal hyperparameters: a penalty parameter $C = 5000$, $\epsilon = 0.1$, and an RBF kernel. The high value of $C$ enables the model to fit complex nonlinear patterns, while $\epsilon$ controls the tolerance for prediction deviations near the regression boundary. 

The XGBoost model was implemented using the \texttt{XGBRegressor} from the \texttt{xgboost} library. XGBoost builds an ensemble of boosted decision trees, where each successive tree is trained to correct the residual errors of the previous ensemble. Grid search with five-fold cross-validation identified the best configuration: 2,000 trees, maximum depth of 9, learning rate of 0.1, subsample ratio of 1.0, and column sampling ratio of 0.6. These settings yield deep, expressive trees while maintaining regularization through feature subsampling. The model objective was set to minimize the squared error loss.

\subsection{Dataset}
In this study, we utilized the AFLOW database, a comprehensive open repository for computational materials science that contains extensive information on inorganic crystalline materials and their properties~\cite{gossett2018aflow, taylor2014restful}. Band gap calculations in AFLOW combine first-principles methods with empirical corrections through an automated workflow. The framework uses VASP to perform DFT calculations with the GGA-PBE functional for standard compounds while applying the GGA+U method for strongly correlated systems containing d- and f-block elements. To address GGA's tendency to underestimate band gaps, AFLOW employs an empirical correction scheme based on a linear regression model derived from 102 benchmark compounds with known experimental values\cite{setyawan2011high,gossett2018aflow,wang2022accurate}. This systematic approach, along with the database's vast size and rich feature space, makes AFLOW particularly well-suited for ML tasks aimed at material property prediction.

For our specific analysis, we selected a subset of 27,600 materials with band gap values ranging between 0 and 5 eV (inclusive). This range was chosen because it encompasses the most relevant band gap values for semiconductors, which are of particular interest in materials science and electronic applications. The lower bound of 0 eV represents materials with metallic behavior, where there is no electronic band gap. Materials that have band gap higher than 5 eV are insulating materials, which have too large band gaps and do not conduct electricity under normal conditions~\cite{tripathy2016optical,masood2023enhancing}. By focusing on materials within this range of 0-5 eV band gaps, we ensure that our model targets materials with practical applications in electronics and optoelectronics.

The dataset was divided into training, validation, and test sets to ensure reliable evaluation and optimization of the model. Specifically, 10\% of the data was reserved for the test set to evaluate the final performance of the model. The remaining 90\% was further split into 80\% for training and 20\% for validation, ensuring sufficient data for model training while retaining a representative validation set. This splitting strategy ensured the distribution of band gap values across all subsets, minimizing sampling bias and enhancing representativeness.

\subsection{Text Data Format}

To investigate the impact of input data representation on model performance, we employed two formats for encoding material property information as text. The first format consists of \texttt{structured strings}, where material attributes, such as chemical composition, crystal structure, and electronic features, were compiled into a consistent, template-based layout. This format emphasizes uniformity and feature alignment across samples, providing a well-controlled structure for the language models to process.

The second format consists of \texttt{natural language descriptions} generated using OpenAI’s GPT-3.5 Turbo API. The same core features were provided to the API to produce narrative-style descriptions, introducing greater linguistic variability and a more conversational tone. Prompts were configured to ensure descriptions remained within a 512-token limit to maintain compatibility with the tokenizer constraints of RoBERTa, and to accommodate input length limits for T5 and Llama-3 as well.

Both formats were applied uniformly across all three models: RoBERTa, T5, Llama-3, and MatSciBERT. The models processed these inputs through their native tokenization pipelines, without additional handcrafted feature engineering. 

%For each model, we appended a custom regression head to enable scalar prediction of band gap values and performed fine-tuning on the downstream task using the corresponding textual inputs. 

\subsection{Input Features}
We carefully selected features that capture both the chemical composition and structural properties of the materials, ensuring a comprehensive understanding of their electronic characteristics, especially the band gap. The selected features include chemical formula, atomic species, valence electron count, crystal symmetry, and magnetic properties, all of which are known to play critical roles in determining the electronic structure and band gap of materials. A complete list of the 23 selected features, categorized by their respective domains, is provided in Table~\ref{tab:feature}.

\begin{table}[htbp]
\centering
\caption{Selected feature list. Each feature is accompanied by a specific description explaining its physical significance and contribution to material characterization.}
\label{tab:feature}
\resizebox{\textwidth}{!}{\begin{tabular}{>{\raggedright\arraybackslash}p{6cm} >{\raggedright\arraybackslash}p{10cm}}
% {\large\begin{tabular}{c p{15cm}p{2cm}}
% {\raggedright\arraybackslash}p{6cm} >{\raggedright\arraybackslash}p{6cm}}
% \begin{tabular}{clcc}
% \hline
\toprule
Feature & Description  \\
\hline
Compound   & Chemical formula of the material, representing its chemical composition \\
Species   & List of atomic species constituting the material  \\
Composition   & Proportion of each element in the material \\
Valence cell (iupac)   & Total number of valence electrons in the unit cell, calculated according to IUPAC standards  \\
Species pseudopotential   & Type of atomic pseudopotentials used for calculations  \\
Crystal class   &  Describing the symmetry properties of the crystal\\
Crystal family   & Indicating the basic geometric features of the crystal \\
Crystal system   & Describing the shape and symmetry of the unit cell \\
Fractional coordinates  & Representing the relative positions of atoms in the unit cell  \\
Lattice parameters   & The edge lengths and angles of the unit cell \\
Lattice system  &  Describing the basic geometric features of the unit cell\\
Lattice variation & Providing a more detailed description of the lattice \\
Space group of the structure   & Describing the symmetry of the crystal  \\
Space group change loose  & Space group determined under looser conditions for crystal structure relaxation, potentially leading to larger symmetry changes \\
Space group change tight  & Space group determined under stricter conditions for structure relaxation, resulting in fewer symmetry changes \\
Point group orbifold  & Describing the topological properties of the point group \\
Point group order  & Indicating the number of symmetry operations in the point group \\
Point group structure  & Describing the geometric features of the point group \\
Point group type  & Classifying the symmetry properties of the point group \\
Magnitude of magnetic moment for each atom   & Describing the local magnetism of the material \\
Magnetization intensity of each atom  & Representing magnetism at the atomic scale \\
Total magnetization intensity of the entire unit cell   & Describing the overall magnetism of the material \\
Density   & Density of the material\\
% \hline
\bottomrule
\end{tabular}
}
\end{table}

The chemical formula represents the basic building blocks of the material, providing critical information about its stoichiometry and composition. The nature and type of atoms constituting the material greatly influence its electronic properties\cite{he2018metallic,zhuo2018predicting,khan2023prediction}. Specifically, the elemental types and their ratios determine atomic energy levels, such as s, p, and d orbitals, as well as bond types like covalent and ionic bonds, and electronegativity differences. These factors affect the relative positions and separation of the valence and conduction bands. In ionic compounds, a larger electronegativity difference leads to a greater energy separation between the valence band, which is primarily formed by anion orbitals, and the conduction band, which is mainly formed by cation orbitals. This typically results in a larger band gap. For example, II-VI compounds often have larger band gaps compared to III-V compounds\cite{wei1998calculated}. For example, the number of valence electrons of each species is crucial for band gap predictions\cite{huang2019band}. The total number of valence electrons per unit cell governs how the electronic bands are filled and directly influences the position of the Fermi level\cite{hu2010modern}. Differences in valence electron count and orbital configuration among elements or compounds can lead to significant variations in both the magnitude and nature of the band gap. Materials with stronger orbital overlap and higher structural symmetry tend to have wider bands and smaller, often direct, band gaps, whereas weaker overlap or lower symmetry can result in indirect or larger gaps, ultimately determining the optical and transport properties of the material\cite{yuan2018unified}.

Additionally, we paid particular attention to structural features, including crystal class, family, and system, as well as lattice parameters, which include the dimensions and angles of the unit cell. These factors not only shape the arrangement of atoms and their interactions but also define the symmetry and geometric properties of the crystal, directly influencing the distribution of electronic states within the energy bands~\cite{zheng2011improving,na2020tuplewise,wang2022accurate}. The crystal's symmetry determines the shape of the Brillouin zone and band degeneracy. High symmetry, as seen in cubic systems, often leads to high band degeneracy at high-symmetry points, where multiple electrons have the same energy, resulting in a relatively simple band structure. In contrast, lower symmetry, as in orthorhombic systems, can cause band splitting, which may change the band gap's size and type\cite{vasseur2013effect}. The space group and point group information were included to account for the effects of symmetry on electronic states and band splitting. Magnetic properties, such as atom magnetic moments and cell magnetization, were also considered due to their relationship with spin distribution and electronegativity\cite{masood2023enhancing}. We specifically chose properties derived from relaxed structures in which the structural configurations have been optimized to minimize energy and stress, ensuring that the atoms are in their equilibrium positions.

\section{Results and Discussion}

\subsection{Framework}
\begin{figure*}[!htb]%[!ht]% %[h!]
\centering
\includegraphics[width=0.9\textwidth]{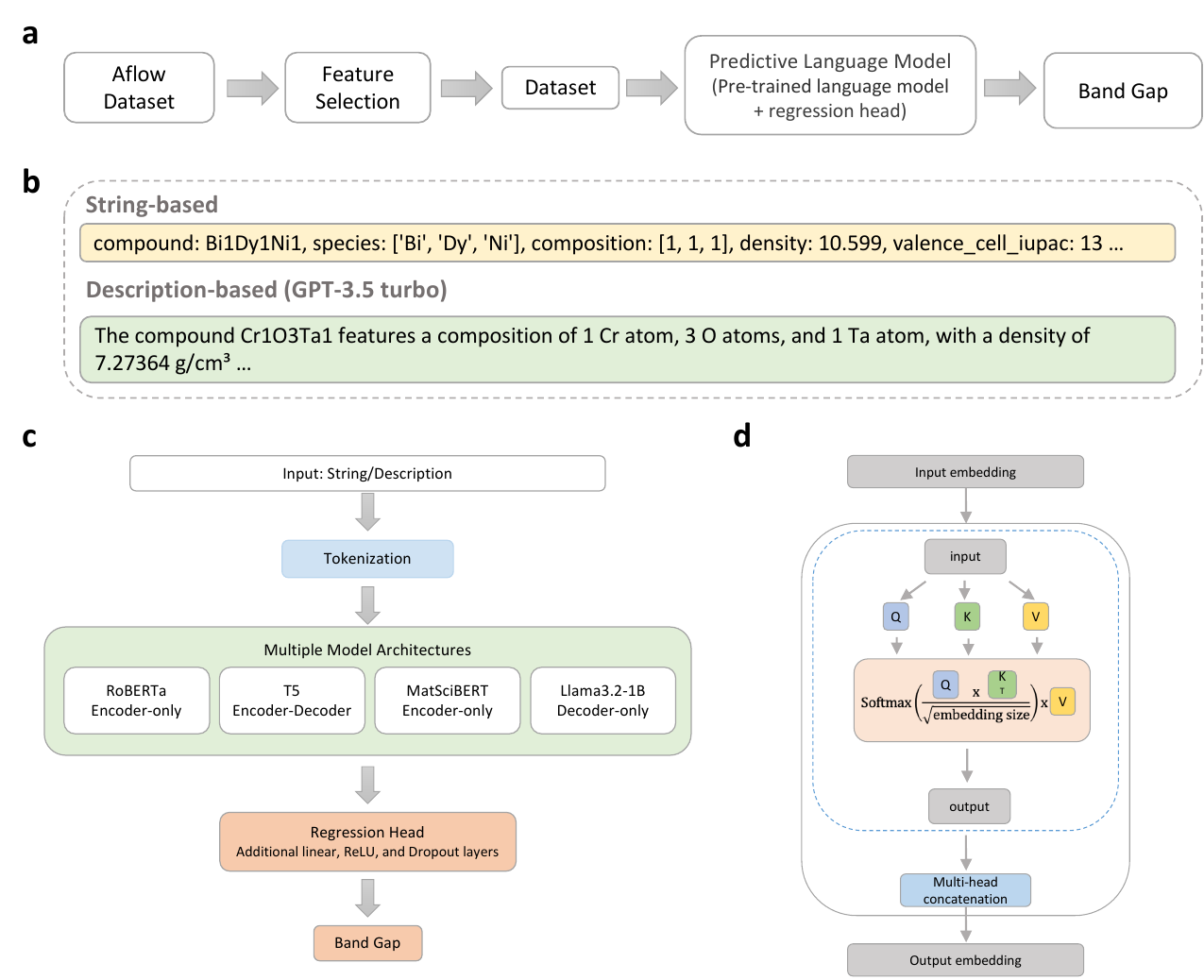} 
\caption{Overview of the proposed band gap prediction framework. \textbf{a} The pipeline starts from the AFLOW dataset, followed by feature selection, dataset preparation, and LLM model training for final band gap prediction. \textbf{b} Two input formats are illustrated. string-based representation using direct feature values and description-based format generated by GPT-3.5 turbo. \textbf{c} Visualization of the finetuning process. The input text undergoes tokenization and embedding through multiple model architectures (RoBERTa, T5, Llama-3, MatSciBERT), followed by a custom regression head for prediction.  \textbf{d} Demonstrates the Transformer encoder and the multi-head attention mechanism with Query (Q), Key (K), and Value (V) operations.}
\label{fig:framework}
\end{figure*}

We developed a language model-based framework to predict the band gaps of semiconductor materials using transformer-based language models: RoBERTa, T5, Llama-3, and MatSciBERT. As shown in Figure~\ref{fig:framework}a, we constructed the dataset from the AFLOW database by extracting relevant material features and transforming them into textual formats suitable for language model inputs. Two types of textual representations were used (Figure~\ref{fig:framework}b): a structured string format that followed a fixed template, and a more flexible natural language description generated using the GPT-3.5 Turbo API. These formats enabled us to assess how each model handles both highly regular and semantically rich input styles. Additional examples of both the structured strings and natural language descriptions are provided in the Supporting Information.

At the core of our approach are transformer-based language models, each differing in architecture, tokenization strategy, and training strategy. Our approach is built on transformer-based language models with distinct architectures and training schemes. RoBERTa and MatSciBERT share an encoder-only design, but MatSciBERT is additionally pretrained on materials science texts for domain-specific understanding. T5 uses an encoder–decoder structure, while Llama-3 is a decoder-only model. Each model tokenizes the input text using its native tokenizer, such as byte-level BPE or SentencePiece, and processes the sequence to generate contextual embeddings (Figure~\ref{fig:framework}c). These embeddings are passed through a custom regression head to produce a scalar band gap prediction. 

%The design of this framework allows for direct comparison across model types and input formats while minimizing the need for handcrafted features or domain-specific encodings.

\subsection{Model Performance}

\begin{figure*}[!htbp]%[htbp] %[h!]
\centering
\includegraphics[width=\textwidth]{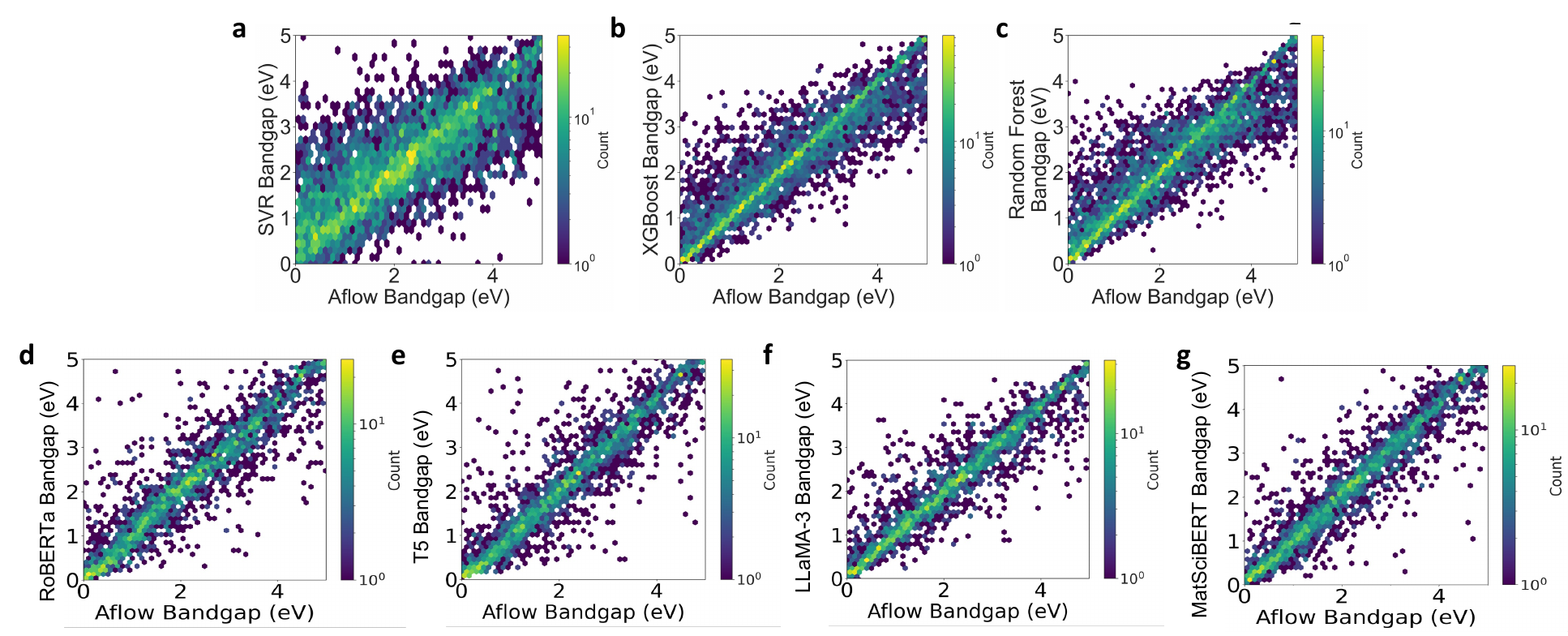} 
\caption{Parity plots for band gap predictions across models: \textbf{a} SVR, \textbf{b} XGBoost, \textbf{c} Random Forest, \textbf{d} RoBERTa, \textbf{e} T5, \textbf{f} Llama-3, \textbf{g} MatSciBERT}
\label{fig:parity}
\end{figure*}

We evaluated the performance of four transformer-based language models—RoBERTa, T5, Llama-3, and MatSciBERT—on the task of predicting semiconductor band gaps from text-based material descriptions. We also included shallow machine learning baselines using the same input format, including SVR, random forest, and XGBoost. GNNs were not considered, as their graph-based input representations do not align with the feature-based approach used here. Model accuracy was quantified using three metrics: mean absolute error (MAE), root mean square error (RMSE), and the coefficient of determination ($R^2$). The results are summarized in Table~\ref{tab:performance}, with parity plots shown in Figure~\ref{fig:parity}.

\begin{table}[h]
\centering
\caption{Comparison of model performance across different ML approaches. For LLMs, results are shown as structured string / natural language description. Best performance per metric is shown in bold, and second-best performance is underlined.}
\label{tab:performance}
\resizebox{\textwidth}{!}{
\begin{tabular}{l l c c c}
\toprule
\textbf{Model} & \textbf{Model Type} & \textbf{MAE (eV)} & \textbf{RMSE (eV)} & \textbf{$R^2$} \\
\midrule
SVR & Shallow ML & $0.601 \pm 0.010$ & $0.844 \pm 0.008$ & $0.600 \pm 0.008$ \\
Random Forest & Shallow ML & $0.385 \pm 0.006$ & $0.609 \pm 0.006$ & $0.792 \pm 0.005$ \\
XGBoost & Shallow ML & $0.318 \pm 0.005$ & $0.537 \pm 0.005$ & $0.838 \pm 0.004$ \\
RoBERTa\textsubscript{(string)} & LLM (Encoder) & $0.325 \pm 0.006$ & $0.447 \pm 0.005$ & $0.855 \pm 0.004$ \\
RoBERTa\textsubscript{(description)} & LLM (Encoder) & $0.421 \pm 0.007$ & $0.590 \pm 0.006$ & $0.797 \pm 0.006$ \\
T5\textsubscript{(string)} & LLM (Encoder-Decoder) & $0.301 \pm 0.007$ & $0.448 \pm 0.006$ & $0.861 \pm 0.005$ \\
T5\textsubscript{(description)} & LLM (Encoder-Decoder) & $0.446 \pm 0.011$ & $0.615 \pm 0.008$ & $0.759 \pm 0.008$ \\
Llama-3\textsubscript{(string)} & LLM (Decoder) & $\mathbf{0.248 \pm 0.006}$ & $\mathbf{0.345 \pm 0.005}$ & $\mathbf{0.891 \pm 0.004}$ \\
Llama-3\textsubscript{(description)} & LLM (Decoder) & $0.335 \pm 0.008$ & $0.473 \pm 0.006$ & $0.843 \pm 0.054$ \\
MatSciBERT\textsubscript{(string)} & LLM (Encoder) &  \underline{$0.288 \pm 0.007$} & \underline{$0.407 \pm 0.006$} & \underline{$0.871 \pm 0.004$} \\
MatSciBERT\textsubscript{(description)} & LLM (Encoder) & $0.366 \pm 0.009$ & $0.503 \pm 0.007$ & $0.808 \pm 0.007$ \\
\bottomrule
\end{tabular}
}
\end{table}

Across all models, inputs in structured string format consistently outperformed descriptive natural language inputs, as shown in Table~\ref{tab:performance}. Both input types are derived from the same underlying features, so the richness of information is equivalent; the difference arises from how the information is presented. Structured strings provide a consistent, systematic representation that reduces variability, making it easier for the model to learn direct mappings between features and target properties. In contrast, language descriptions introduce variability in phrasing, terminology, and length, requiring the model to first infer the relevant features from context. This adds complexity and can obscure patterns, even though the same information is present.

Among the transformer-based models, Llama-3 with structured input achieved the best overall performance, with a mean absolute error of 0.248 eV and $R^2$ of 0.891 (Table~\ref{tab:performance}). Following closely, MatSciBERT delivered a strong performance with an MAE of 0.288 eV and an $R^2$
of 0.872. While RoBERTa and T5 also outperformed the shallow ML models, their performance was lower when using GPT-generated descriptions compared to structured string inputs.

These results highlight several insights. First, language models can predict band gaps directly from text-based inputs without any feature preprocessing. Given its large number of parameters, it is not surprising that Llama-3 performed the best. More interestingly, when comparing RoBERTa and MatSciBERT, which share similar architectures, MatSciBERT achieved better results despite having fewer parameters. This suggests that the training strategy and the use of a domain-specific corpus are critical. Pretraining on materials science literature enables the model to more effectively capture the relationship between material features and band gap values.

Even though we evaluated multiple types of language models, it is difficult to generalize performance purely based on architecture since the number of parameters varies greatly. Nonetheless, it is important to demonstrate that text-based band gap prediction works across different model types. It is also not surprising that language model approaches outperform shallow machine learning methods. To clarify, both shallow ML models and language models use the same feature set, so the richness of information is identical. The superior performance of language models indicates their enhanced capability to learn complex correlations between the provided features and the band gap.

% This result can be attributed to the inherent differences in information density and noise. Structured strings provide a consistent, low-noise representation where the position and syntax of each feature are predictable and semantically aligned. The models can more easily learn a direct mapping between these consistent textual cues and the target property. In contrast, while the GPT-generated descriptions are linguistically rich, they introduce variability and language noise that can obscure the direct link between specific material features and their corresponding numerical values, even for powerful models like LLaMA-3. However, the ability of all four language models to still learn meaningful correlations from unstructured natural language input underscores a key strength of this approach. Rather than viewing this variability as a limitation, we interpret it as evidence of the models’ capacity to generalize from flexible, human-like descriptions of materials. Notably, even in its smallest size, LLaMA-3 demonstrated strong capacity for property prediction when paired with fine-tuning, indicating that decoder-only architectures can serve as lightweight yet powerful alternatives for scientific regression tasks.

\FloatBarrier 
\subsection{Layer Freezing Analysis}

To investigate how pretrained representations in each model contribute to band gap prediction, we conducted layer-freezing experiments on RoBERTa, T5, Llama-3, and MatSciBERT using structured string inputs. In these experiments, we progressively froze layers of the transformer backbone and fine-tuned only a portion of the model, while training a custom regression head on top of the frozen or partially frozen representations. The goal was not to reduce computational cost, but to probe which layers carry the most task-relevant information. Table~\ref{tab:freeze_combined} reports the MAE, RMSE, and $R^2$ values for each model under various freezing setups. This analysis provides insight into how much domain-specific fine-tuning is necessary and which parts of the pretrained model are most informative for the downstream task, highlighting the layers that contribute most strongly to predictive accuracy.

\begin{table}[hbt]
\centering
\caption{Comparison of layer freezing strategies across RoBERTa, T5, Llama-3, and MatSciBERT using structured string inputs. The fully finetuned (non-frozen) results are included for reference. Parameter counts and percentages represent the number of trainable parameters relative to the no-freezing case. All values are reported as mean $\pm$ standard deviation.}
\label{tab:freeze_combined}
\resizebox{\textwidth}{!}{
\begin{tabular}{llcccc}
\toprule
\textbf{Model} & \textbf{Freezing Strategy} & \textbf{Parameters (count, \%)} & \textbf{MAE (eV)} & \textbf{RMSE (eV)} & \textbf{$R^2$} \\
\midrule
\multirow{5}{*}{RoBERTa} 
  & Fully finetuned (no freezing)     & 125,236,993 (100\%) & $0.325 \pm 0.006$ & $0.447 \pm 0.005$ & $0.855 \pm 0.004$ \\
  & Freeze first layer                 & 118,149,121 (94.3\%) & $0.328 \pm 0.008$ & $0.448 \pm 0.006$ & $0.848 \pm 0.005$ \\
  & Freeze all but final 3 layers      & 61,446,145 (49.1\%) & $0.388 \pm 0.009$ & $0.510 \pm 0.007$ & $0.817 \pm 0.006$ \\
  & Freeze all but final layer         & 8,269,825 (6.6\%) & $0.509 \pm 0.012$ & $0.648 \pm 0.009$ & $0.721 \pm 0.009$ \\
  & Freeze all layers                  & 591,361 (0.5\%) & $0.663 \pm 0.016$ & $0.826 \pm 0.011$ & $0.563 \pm 0.013$ \\
\midrule
\multirow{5}{*}{T5} 
  & Fully finetuned (no freezing)     & 60,769,793 (100\%) & $0.301 \pm 0.007$ & $0.448 \pm 0.006$ & $0.861 \pm 0.005$ \\
  & Freeze first layer                 & 57,622,785 (94.8\%) & $0.350 \pm 0.008$ & $0.504 \pm 0.007$ & $0.849 \pm 0.006$ \\
  & Freeze all but final 3 layers      & 21,830,401 (19.8\%) & $0.367 \pm 0.009$ & $0.516 \pm 0.007$ & $0.832 \pm 0.006$ \\
  & Freeze all but final layer         & 7,671,553 (7.0\%) & $0.598 \pm 0.014$ & $0.784 \pm 0.011$ & $0.619 \pm 0.011$ \\
  & Freeze all layers                  & 592,129 (0.5\%) & $0.792 \pm 0.019$ & $0.981 \pm 0.013$ & $0.420 \pm 0.014$ \\
\midrule
\multirow{5}{*}{Llama-3} 
  & Fully finetuned (no freezing)     & 1,237,915,649 (100\%) & $\mathbf{0.248 \pm 0.006}$ & $\mathbf{0.345 \pm 0.005}$ & $\mathbf{0.891 \pm 0.004}$ \\
  & Freeze first layer                 & 1,177,094,145 (95.1\%) & $0.279 \pm 0.007$ & $0.426 \pm 0.006$ & $0.878 \pm 0.004$ \\
  & Freeze all but final 3 layers      & 447,236,097 (36.1\%) & $0.318 \pm 0.008$ & $0.474 \pm 0.006$ & $0.851 \pm 0.005$ \\
  & Freeze all but final layer         & 325,593,089 (26.3\%) & $0.424 \pm 0.010$ & $0.576 \pm 0.008$ & $0.793 \pm 0.007$ \\
  & Freeze all layers                  & 2,101,249 (0.2\%) & $0.716 \pm 0.017$ & $0.893 \pm 0.012$ & $0.518 \pm 0.013$ \\
\midrule
\multirow{5}{*}{MatSciBERT} 
  & Fully finetuned (no freezing)     & 110,509,825 (100\%) & $0.288 \pm 0.007$ & $0.407 \pm 0.006$ & $0.871 \pm 0.004$ \\
  & Freeze first layer                 & 103,421,953 (93.6\%) & $0.293 \pm 0.007$ & $0.404 \pm 0.005$ & $0.874 \pm 0.004$ \\
  & Freeze all but final 3 layers      & 22,445,569 (20.3\%) & $0.340 \pm 0.008$ & $0.454 \pm 0.006$ & $0.849 \pm 0.005$ \\
  & Freeze all but final layer         & 8,269,825 (7.5\%) & $0.415 \pm 0.010$ & $0.539 \pm 0.007$ & $0.798 \pm 0.007$ \\
  & Freeze all layers                  & 1,181,953 (1.1\%) & $0.827 \pm 0.020$ & $0.994 \pm 0.013$ & $0.385 \pm 0.015$ \\
\bottomrule
\end{tabular}
}
\end{table}

Throughout this study, we refer to the original language models, trained on general text corpora prior to any materials-specific adaptation, as ``pretrained" models. After supervised training on the band gap prediction task, we refer to the models as ``finetuned."

% In all four models, predictive performance improved progressively as more transformer layers were unfrozen, highlighting the importance of task-specific fine-tuning in scientific regression applications. This trend highlights a key insight: while LLMs capture broadly useful representations during pretraining, these representations alone are insufficient to achieve optimal accuracy in domain-specific prediction tasks without additional adaptation. In the case of RoBERTa and T5, the most notable performance gains occurred when the final three layers of the model were unfrozen. This suggests that the highest-level contextual embeddings learned during pretraining are not directly aligned with the band gap target property but can be effectively adapted through partial retraining.

In all four models, predictive performance improved progressively as more transformer layers were unfrozen, highlighting the importance of task-specific fine-tuning in scientific regression tasks, as shown in Table~\ref{tab:freeze_combined}. While LLMs capture broadly useful representations during pretraining, these representations alone are insufficient to achieve optimal accuracy in domain-specific prediction tasks without additional adaptation. As more layers are involved in fine-tuning, the models can better capture feature–property correlations. Interestingly, MatSciBERT exhibits slightly better RMSE and $R^2$ scores compared to its fully finetuned case, which may be attributed to the well-formed representations learned from prior exposure to the materials science domain.

\begin{figure*}[!htb]
\centering
\includegraphics[width=0.7\textwidth]{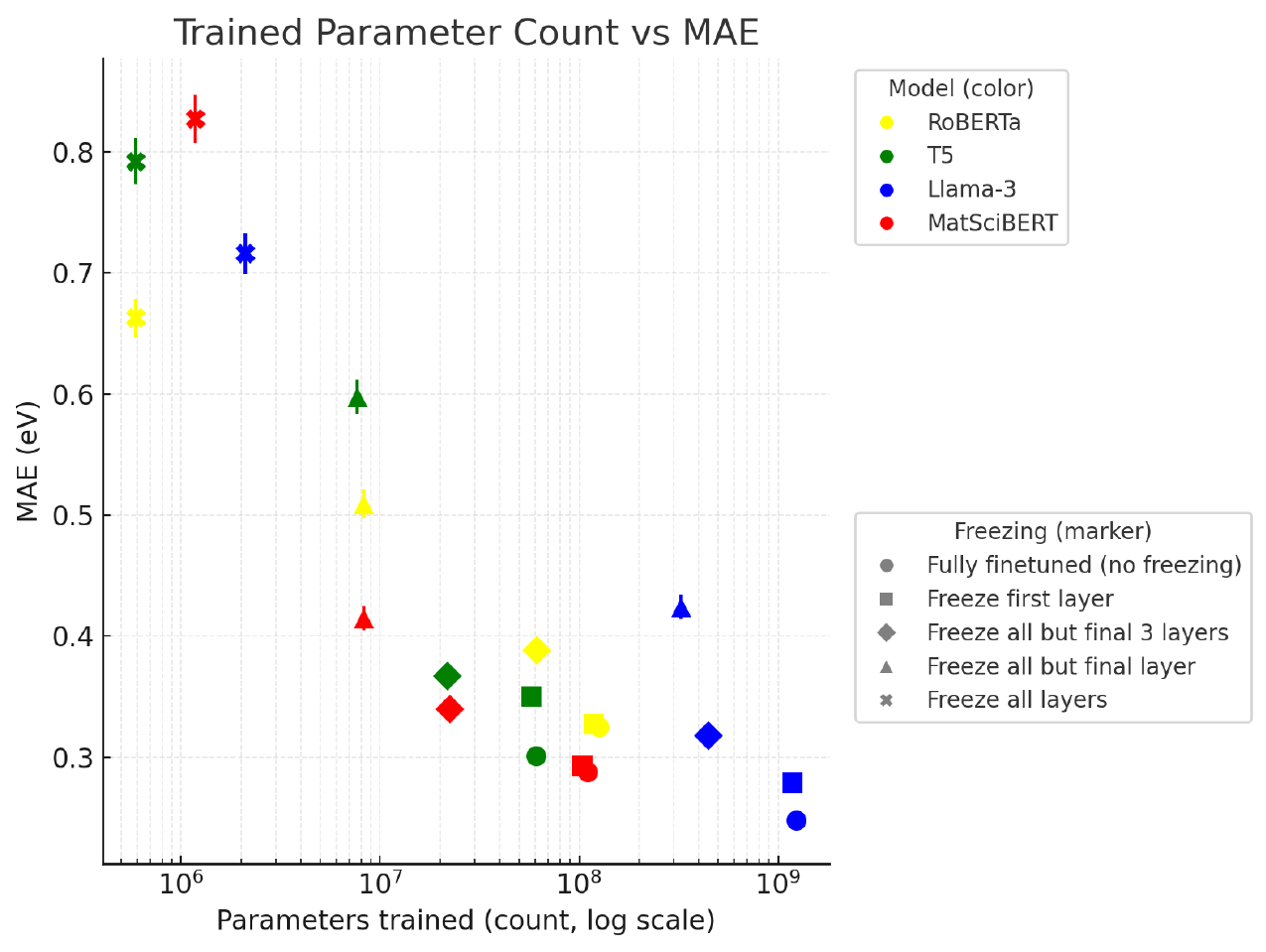}
\caption{Scaling behavior of finetuning strategies across transformer-based models. MAE is shown as a function of the number of trainable parameters. Colors indicate different model architectures: RoBERTa (yellow), T5 (green), Llama-3 (blue), and MatSciBERT (red). Marker shapes represent different freezing strategies, from fully finetuned (no freezing) to full layer freezing.}
\label{fig:scalingplot}
\end{figure*}

The rough scaling relationship between the number of parameters and prediction accuracy is illustrated in Figure~\ref{fig:scalingplot}. Generally, as the number of trainable parameters increases, MAE decreases, indicating improved predictive performance across models and layer-freezing strategies. Within each model, this relationship is relatively linear in the layer-wise freezing experiments, showing that gradually unfreezing layers consistently improves performance. However, when comparing across different models, the relationship is not strictly linear. This reflects the fact that predictive performance depends not only on the number of parameters but also on other factors, including model architecture, pretraining corpus, task-specific fine-tuning strategies, and the alignment between learned representations and the structured input features. These results emphasize that while pretrained LLMs provide a strong foundation, achieving optimal performance in materials science requires careful selection of trainable layers and fine-tuning strategies, especially under computational or resource constraints.

\subsection{Feature-wise Self-Attention Score}

To identify which material features are most emphasized during band gap prediction, we conducted a feature-wise self-attention analysis on Llama-3 and MatSciBERT, the two best-performing models. In transformer architectures, the attention mechanism quantifies how strongly each token attends to others when forming contextual representations, as shown in Figure~\ref{fig:framework}d.

For each model, input text was tokenized (maximum length of 512 tokens, truncation enabled), and attention weights were extracted from the first and last transformer layers. The attention score here refers to the attention weight between the first token, used as the regression header, and all other tokens in the sequence. For each head, the attention scores from the regression token to the rest of the tokens were averaged, and the resulting values were then averaged across all heads within the layer.

Attention scores between the regression header token and all feature tokens were extracted from each transformer layer. 
For a given layer \( l \) and sample \( i \), the attention weights were first averaged across all attention heads:
\begin{equation}
s_{i,l}(t) = \frac{1}{H} \sum_{h=1}^{H} s_{i,l,h}(t)
\end{equation}
where \( H \) is the number of attention heads and \( s_{i,l,h}(t) \) denotes the attention score from the header token to token \( t \) in head \( h \).

For each feature \( f \), we identified its corresponding token span and recorded the maximum attention score within that span:
\begin{equation}
\text{score}_{i,l,f}^{\text{raw}} = \max_{t \in [t_{\text{start}}, t_{\text{end}}]} s_{i,l}(t)
\end{equation}

To ensure comparability across samples, raw scores were normalized to the range [0, 1] using min--max scaling:
\begin{equation}
\text{score}_{i,l,f}^{\text{norm}} = 
\frac{\text{score}_{i,l,f}^{\text{raw}} - \min_k \text{score}_{i,l,k}^{\text{raw}}}
{\max_k \text{score}_{i,l,k}^{\text{raw}} - \min_k \text{score}_{i,l,k}^{\text{raw}}}
\end{equation}
where \( k \) indexes all features in sample \( i \).

Finally, layer-wise feature attention was obtained by averaging normalized scores across all test samples:
\begin{equation}
\text{Avg\_Attention}_{l,f} = \frac{1}{N} \sum_{i=1}^{N} \text{score}_{i,l,f}^{\text{norm}}
\end{equation}

This procedure provides a consistent measure of how each layer attends to specific material features based on attention scores. It is important to note that these scores do not incorporate the value matrix; therefore, as illustrated in Figure~\ref{fig:framework}d, they do not directly quantify feature importance. Instead, they offer a descriptive view of the model’s focus on the input tokens.

\begin{figure*}[!htb]
\centering
\includegraphics[width=\textwidth]{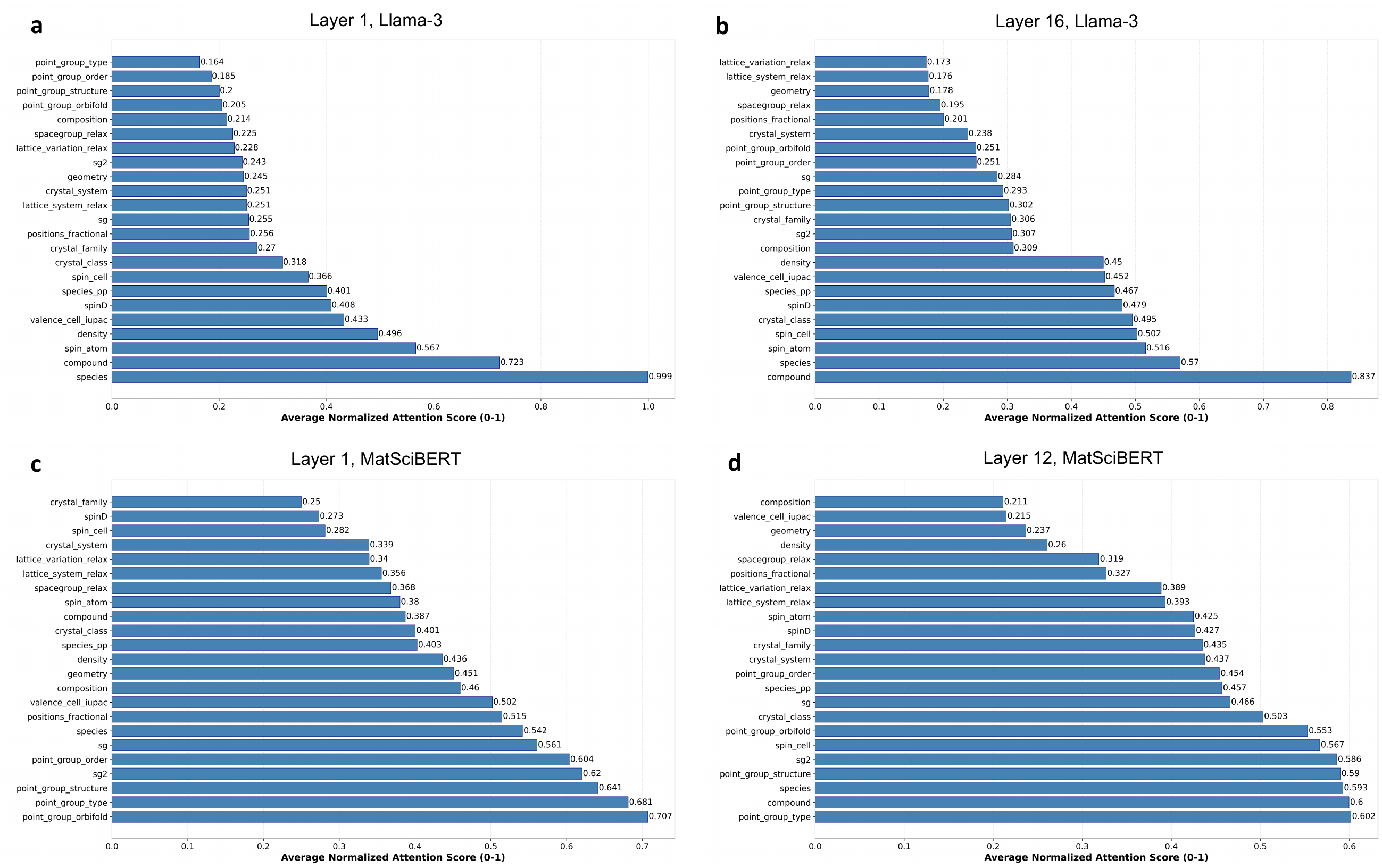} 
\caption{Feature-wise self-attention scores for LLaMA-3 and MatSciBERT. \textbf{a} LLaMA-3, first layer; \textbf{b} LLaMA-3, final layer (layer 16); \textbf{c} MatSciBERT, first layer; \textbf{d} MatSciBERT, final layer (layer 12).}
\label{fig:attention}
\end{figure*}

The results reveal clear differences in attention patterns between the first and last layers for both models. In our analysis, the embedding of the first token from the last layer is ultimately used for regression, reflecting how the model allocates attention to optimize prediction performance. For LLaMA-3, attention to geometric features, such as lattice variation, lattice system, and overall geometry, decreases in the final layer, as seen by comparing Figure~\ref{fig:attention}a and b. Although these features are important for structure–property relationships, this pattern suggests that LLaMA-3 struggles to directly extract geometric information from text-based descriptions, instead emphasizing compositional and spin-related features.

MatSciBERT, which benefits from domain-specific pretraining on materials science literature, shows earlier specialization in crystallographic features, as illustrated in Figure~\ref{fig:attention}c. Similar to LLaMA-3, attention to geometric features declines in later layers, highlighting the difficulty of encoding geometric information purely from text (Figure~\ref{fig:attention}d). However, MatSciBERT distributes attention more broadly in the first layer, with structural symmetry attributes such as point group receiving relatively high scores. By the final layer, attention becomes more selective, focusing on key features such as composition, including species and compound information.

\subsection{Embedding Map}
\begin{figure*}[!htb] %[h!]
\centering
\makebox[\textwidth][c]{%
  \includegraphics[width=1.1\textwidth]{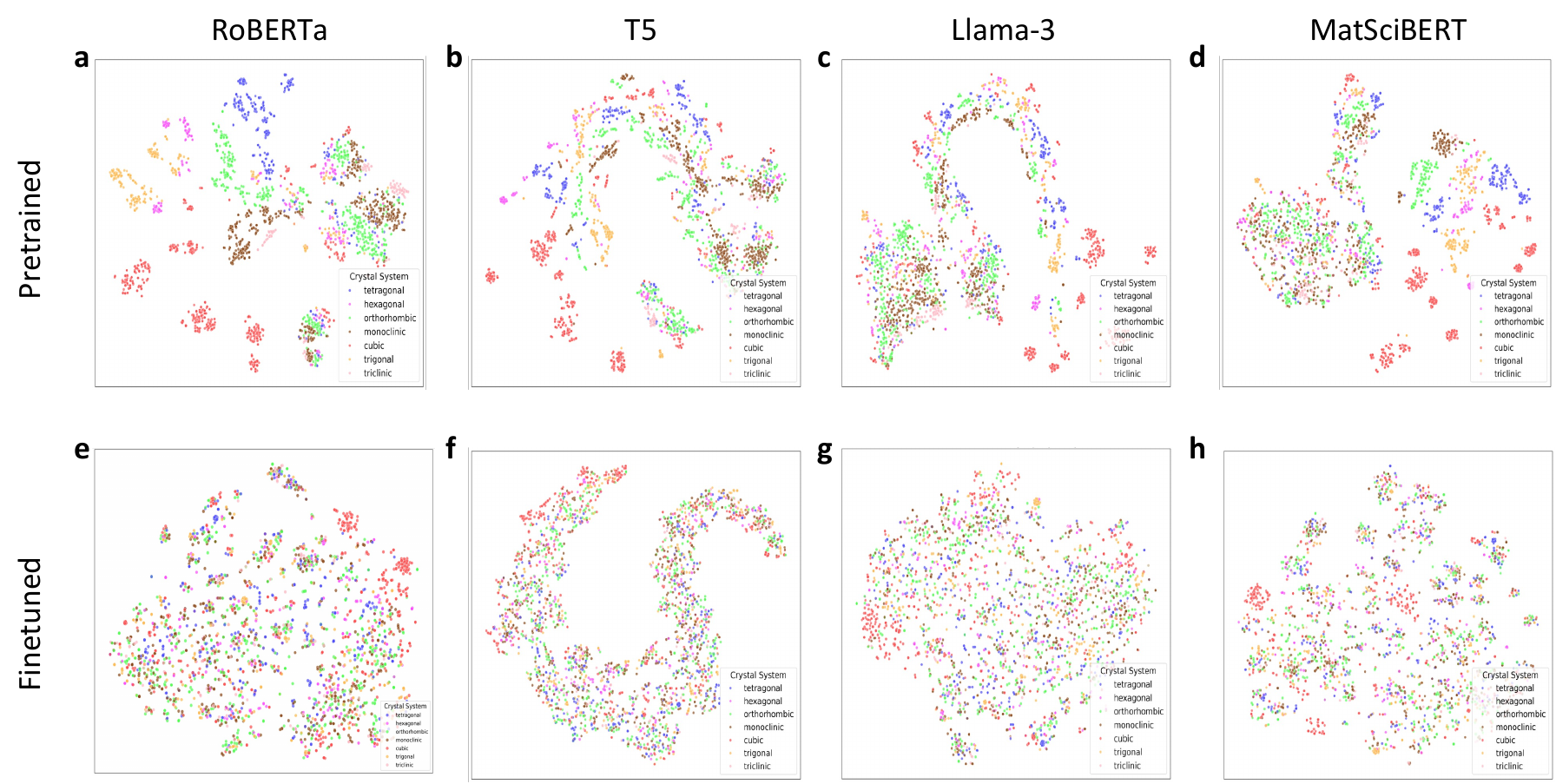}
}
\caption{t-SNE visualizations of embeddings colored by crystal system. \textbf{a-d} show results from the pretrained models: \textbf{a} RoBERTa, \textbf{b} T5, \textbf{c} Llama-3, \textbf{d} MatSciBERT. \textbf{e-h} show results from the corresponding finetuned models: \textbf{e} RoBERTa, \textbf{f} T5, \textbf{g} Llama-3, \textbf{h} MatSciBERT.}
\label{fig:tSNE_crystal}
\end{figure*}

\begin{figure*}[!htb] %[h!]
\centering
\makebox[\textwidth][c]{%
  \includegraphics[width=1.1\textwidth]{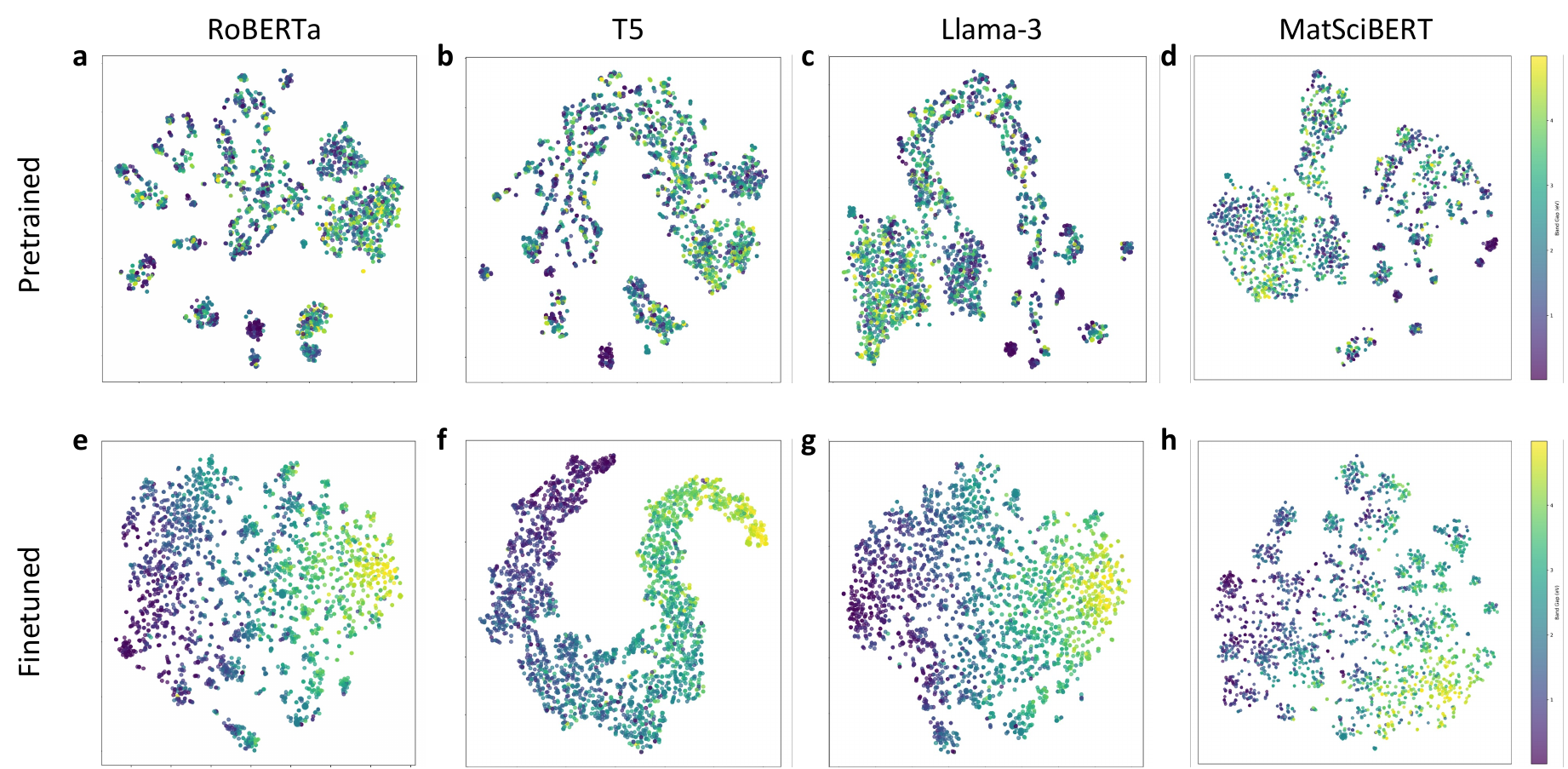}
}
\caption{t-SNE visualizations of embeddings for bandgap prediction. \textbf{a-d} show results from the pretrained models: \textbf{a} RoBERTa, \textbf{b} T5, \textbf{c} Llama-3, \textbf{d} MatSciBERT. \textbf{e-h} show results from the corresponding finetuned models: \textbf{e} RoBERTa, \textbf{f} T5, \textbf{g} Llama-3, \textbf{h} MatSciBERT.}
\label{fig:tSNE_bandgap}
\end{figure*}

We conducted embedding space analysis using t-SNE visualizations to investigate how each model organizes material representations before and after fine-tuning. As shown in Figure~\ref{fig:tSNE_crystal} and \ref{fig:tSNE_bandgap}, we extracted the first-token embeddings from the pretrained and finetuned versions of RoBERTa, T5, Llama-3, and MatSciBERT using structured string inputs, and colored the points by crystal system. In the ``pretrained" condition, which corresponds to the original language model weights before finetuning on the band gap prediction task, the embeddings reflect structural signals learned from general language corpora. The ``fine-tuned" condition represents embeddings generated after the models were fully fine-tuned for band gap prediction.

In the pretrained state, embeddings from all four models exhibit meaningful clustering according to crystal system, as shown in Figure~\ref{fig:tSNE_crystal}a-d. In contrast, when coloring by bandgap (Figure~\ref{fig:tSNE_bandgap}a-d), no clear clustering is observed. This structure-aware behavior likely arises from explicit input features such as crystal class, space group, and lattice parameters, which correlate with symmetry. While these features can separate materials by structural family, they do not provide sufficient signal to predict electronic properties like bandgap. In other words, without finetuning, the models lack task-specific supervision to map structural cues to quantitative property outcomes.

After finetuning on the bandgap prediction task, however, the clustering behavior changes markedly: embeddings now group primarily by bandgap rather than by crystal system, as visualized in Figure~\ref{fig:tSNE_crystal}e-h and Figure~\ref{fig:tSNE_bandgap}e-h. In particular, Llama-3 shows a significant loss of crystal system clustering while forming a near-linear embedding map based on bandgap, which likely contributes to its superior predictive performance. The result is an embedding space where materials with similar bandgaps are positioned closer together, independent of their crystallographic classification. This transformation illustrates how supervised training can reorient general-purpose language representations toward property-specific scientific objectives.

T5 produces a qualitatively different t-SNE embedding: unlike the roughly circular maps seen in most other models, T5’s embedding forms a curvy, elongated shape. We note that the models have differing embedding dimensions, and t-SNE can distort distances depending on dimensionality and distribution. Therefore, some apparent differences may reflect visualization artifacts or latent space geometry rather than intrinsic model behavior. Further quantitative analysis of cluster separation and latent structure would be needed to confirm these patterns.

\section{Conclusion}

In this study, we investigated the use of transformer-based language models, RoBERTa, T5, Llama-3, and MatSciBERT for predicting the band gaps of semiconductor materials directly from textual inputs. We demonstrated that these models can learn meaningful feature–property relationships without relying on complex feature preprocessing or structure graph of atomic coordinates. Llama-3 achieving the highest accuracy using structured inputs (MAE 0.248~eV, $R^2$ 0.891). Even with natural language descriptions, the models captured relevant patterns, highlighting their flexibility for scenarios lacking structured data.

Overall, this work highlights the potential of both general-purpose and domain-specific language models as flexible, scalable, and efficient tools for materials property prediction. By enabling direct prediction of properties from human-readable text, these models remove the need for extensive feature engineering or graph-based structure encoding, allowing rapid, end-to-end property estimation from structured or natural language material descriptions. The results also demonstrate that pretrained language models can provide interpretable insights through attention and embedding analyses, identifying the material features most relevant to the target property. .
% Overall, our results establish transformer-based language models, particularly compact, decoder-only architectures like Llama-3, as promising tools for text-driven materials property prediction, offering scalability, interpretability, and efficiency for future materials informatics pipelines.



\section*{Code Availability Statement}
The Python code in this study is available on GitHub at the following link: \url{https://github.com/yingtiny/bandgap_prediction_RoBERTa}.

\section*{Technology Use Disclosure}
We used ChatGPT and Claude to help with grammar and typographical corrections during the preparation of this manuscript. The authors have carefully reviewed, verified, and approved all content to ensure accuracy and integrity.
%\url{}.
\bibliography{reference}

\providecommand{\latin}[1]{#1}
\makeatletter
\providecommand{\doi}
  {\begingroup\let\do\@makeother\dospecials
  \catcode`\{=1 \catcode`\}=2 \doi@aux}
\providecommand{\doi@aux}[1]{\endgroup\texttt{#1}}
\makeatother
\providecommand*\mcitethebibliography{\thebibliography}
\csname @ifundefined\endcsname{endmcitethebibliography}  {\let\endmcitethebibliography\endthebibliography}{}
\begin{mcitethebibliography}{39}
\providecommand*\natexlab[1]{#1}
\providecommand*\mciteSetBstSublistMode[1]{}
\providecommand*\mciteSetBstMaxWidthForm[2]{}
\providecommand*\mciteBstWouldAddEndPuncttrue
  {\def\EndOfBibitem{\unskip.}}
\providecommand*\mciteBstWouldAddEndPunctfalse
  {\let\EndOfBibitem\relax}
\providecommand*\mciteSetBstMidEndSepPunct[3]{}
\providecommand*\mciteSetBstSublistLabelBeginEnd[3]{}
\providecommand*\EndOfBibitem{}
\mciteSetBstSublistMode{f}
\mciteSetBstMaxWidthForm{subitem}{(\alph{mcitesubitemcount})}
\mciteSetBstSublistLabelBeginEnd
  {\mcitemaxwidthsubitemform\space}
  {\relax}
  {\relax}

\bibitem[Yu and Cardona(2010)Yu, and Cardona]{yu2010fundamentals}
Yu,~P.~Y.; Cardona,~M. \emph{Fundamentals of Semiconductors: Physics and Materials Properties}, 4th ed.; Springer, 2010\relax
\mciteBstWouldAddEndPuncttrue
\mciteSetBstMidEndSepPunct{\mcitedefaultmidpunct}
{\mcitedefaultendpunct}{\mcitedefaultseppunct}\relax
\EndOfBibitem
\bibitem[Kim \latin{et~al.}(2020)Kim, Lee, Hong, Yoon, An, Lee, Jeong, Yoo, Kang, Youn, \latin{et~al.} others]{kim2020band}
Kim,~S.; Lee,~M.; Hong,~C.; Yoon,~Y.; An,~H.; Lee,~D.; Jeong,~W.; Yoo,~D.; Kang,~Y.; Youn,~Y.; others A band-gap database for semiconducting inorganic materials calculated with hybrid functional. \emph{Scientific Data} \textbf{2020}, \emph{7}, 387\relax
\mciteBstWouldAddEndPuncttrue
\mciteSetBstMidEndSepPunct{\mcitedefaultmidpunct}
{\mcitedefaultendpunct}{\mcitedefaultseppunct}\relax
\EndOfBibitem
\bibitem[Masood \latin{et~al.}(2023)Masood, Sirojan, Toe, Kumar, Haghshenas, Sit, Amal, Sethu, and Teoh]{masood2023enhancing}
Masood,~H.; Sirojan,~T.; Toe,~C.~Y.; Kumar,~P.~V.; Haghshenas,~Y.; Sit,~P.~H.; Amal,~R.; Sethu,~V.; Teoh,~W.~Y. Enhancing prediction accuracy of physical band gaps in semiconductor materials. \emph{Cell Reports Physical Science} \textbf{2023}, \emph{4}\relax
\mciteBstWouldAddEndPuncttrue
\mciteSetBstMidEndSepPunct{\mcitedefaultmidpunct}
{\mcitedefaultendpunct}{\mcitedefaultseppunct}\relax
\EndOfBibitem
\bibitem[Koch and Holthausen(2015)Koch, and Holthausen]{koch2015chemist}
Koch,~W.; Holthausen,~M.~C. \emph{A chemist's guide to density functional theory}; John Wiley \& Sons, 2015\relax
\mciteBstWouldAddEndPuncttrue
\mciteSetBstMidEndSepPunct{\mcitedefaultmidpunct}
{\mcitedefaultendpunct}{\mcitedefaultseppunct}\relax
\EndOfBibitem
\bibitem[Kohn and Sham(1965)Kohn, and Sham]{kohn1965self}
Kohn,~W.; Sham,~L.~J. Self-consistent equations including exchange and correlation effects. \emph{Physical review} \textbf{1965}, \emph{140}, A1133\relax
\mciteBstWouldAddEndPuncttrue
\mciteSetBstMidEndSepPunct{\mcitedefaultmidpunct}
{\mcitedefaultendpunct}{\mcitedefaultseppunct}\relax
\EndOfBibitem
\bibitem[Schleder \latin{et~al.}(2019)Schleder, Padilha, Acosta, Costa, and Fazzio]{schleder2019dft}
Schleder,~G.~R.; Padilha,~A.~C.; Acosta,~C.~M.; Costa,~M.; Fazzio,~A. From DFT to machine learning: recent approaches to materials science--a review. \emph{Journal of Physics: Materials} \textbf{2019}, \emph{2}, 032001\relax
\mciteBstWouldAddEndPuncttrue
\mciteSetBstMidEndSepPunct{\mcitedefaultmidpunct}
{\mcitedefaultendpunct}{\mcitedefaultseppunct}\relax
\EndOfBibitem
\bibitem[Wang \latin{et~al.}(2021)Wang, Tan, Wei, and Jin]{wang2021accurate}
Wang,~T.; Tan,~X.; Wei,~Y.; Jin,~H. Accurate bandgap predictions of solids assisted by machine learning. \emph{Materials Today Communications} \textbf{2021}, \emph{29}, 102932\relax
\mciteBstWouldAddEndPuncttrue
\mciteSetBstMidEndSepPunct{\mcitedefaultmidpunct}
{\mcitedefaultendpunct}{\mcitedefaultseppunct}\relax
\EndOfBibitem
\bibitem[Rajan \latin{et~al.}(2018)Rajan, Mishra, Satsangi, Vaish, Mizuseki, Lee, and Singh]{rajan2018machine}
Rajan,~A.~C.; Mishra,~A.; Satsangi,~S.; Vaish,~R.; Mizuseki,~H.; Lee,~K.-R.; Singh,~A.~K. Machine-learning-assisted accurate band gap predictions of functionalized MXene. \emph{Chemistry of Materials} \textbf{2018}, \emph{30}, 4031--4038\relax
\mciteBstWouldAddEndPuncttrue
\mciteSetBstMidEndSepPunct{\mcitedefaultmidpunct}
{\mcitedefaultendpunct}{\mcitedefaultseppunct}\relax
\EndOfBibitem
\bibitem[Zhuo \latin{et~al.}(2018)Zhuo, Mansouri~Tehrani, and Brgoch]{zhuo2018predicting}
Zhuo,~Y.; Mansouri~Tehrani,~A.; Brgoch,~J. Predicting the band gaps of inorganic solids by machine learning. \emph{The journal of physical chemistry letters} \textbf{2018}, \emph{9}, 1668--1673\relax
\mciteBstWouldAddEndPuncttrue
\mciteSetBstMidEndSepPunct{\mcitedefaultmidpunct}
{\mcitedefaultendpunct}{\mcitedefaultseppunct}\relax
\EndOfBibitem
\bibitem[Faber \latin{et~al.}(2017)Faber, Hutchison, Huang, and von Lilienfeld]{faber2017prediction}
Faber,~F.~A.; Hutchison,~G.~R.; Huang,~B.; von Lilienfeld,~O.~A. Prediction errors of molecular machine learning models lower than hybrid DFT error. \emph{Journal of Chemical Theory and Computation} \textbf{2017}, \emph{13}, 5255--5264\relax
\mciteBstWouldAddEndPuncttrue
\mciteSetBstMidEndSepPunct{\mcitedefaultmidpunct}
{\mcitedefaultendpunct}{\mcitedefaultseppunct}\relax
\EndOfBibitem
\bibitem[Choudhary and DeCost(2021)Choudhary, and DeCost]{choudhary2021atomistic}
Choudhary,~K.; DeCost,~B. The Atomistic Line Graph Neural Network for improved materials property predictions. \emph{npj Computational Materials} \textbf{2021}, \emph{7}, 185\relax
\mciteBstWouldAddEndPuncttrue
\mciteSetBstMidEndSepPunct{\mcitedefaultmidpunct}
{\mcitedefaultendpunct}{\mcitedefaultseppunct}\relax
\EndOfBibitem
\bibitem[Taniguchi \latin{et~al.}(2023)Taniguchi, Hosokawa, and Asahi]{taniguchi2023graph}
Taniguchi,~T.; Hosokawa,~M.; Asahi,~T. Graph comparison of molecular crystals in band gap prediction using neural networks. \emph{ACS omega} \textbf{2023}, \emph{8}, 39481--39489\relax
\mciteBstWouldAddEndPuncttrue
\mciteSetBstMidEndSepPunct{\mcitedefaultmidpunct}
{\mcitedefaultendpunct}{\mcitedefaultseppunct}\relax
\EndOfBibitem
\bibitem[Ock \latin{et~al.}(2023)Ock, Guntuboina, and Barati~Farimani]{ock2023catberta}
Ock,~J.; Guntuboina,~C.; Barati~Farimani,~A. Catalyst Energy Prediction with CatBERTa: Unveiling Feature Exploration Strategies through Large Language Models. \emph{ACS Catalysis} \textbf{2023}, \emph{13}, 16032--16044\relax
\mciteBstWouldAddEndPuncttrue
\mciteSetBstMidEndSepPunct{\mcitedefaultmidpunct}
{\mcitedefaultendpunct}{\mcitedefaultseppunct}\relax
\EndOfBibitem
\bibitem[Ock \latin{et~al.}(2024)Ock, Badrinarayanan, Magar, Antony, and Barati~Farimani]{ock2024multimodal}
Ock,~J.; Badrinarayanan,~S.; Magar,~R.; Antony,~A.; Barati~Farimani,~A. Multimodal language and graph learning of adsorption configuration in catalysis. \emph{Nature Machine Intelligence} \textbf{2024}, 1--11\relax
\mciteBstWouldAddEndPuncttrue
\mciteSetBstMidEndSepPunct{\mcitedefaultmidpunct}
{\mcitedefaultendpunct}{\mcitedefaultseppunct}\relax
\EndOfBibitem
\bibitem[Guntuboina \latin{et~al.}(2023)Guntuboina, Das, Mollaei, Kim, and Barati~Farimani]{peptidebert}
Guntuboina,~C.; Das,~A.; Mollaei,~P.; Kim,~S.; Barati~Farimani,~A. PeptideBERT: A Language Model Based on Transformers for Peptide Property Prediction. \emph{The Journal of Physical Chemistry Letters} \textbf{2023}, \emph{14}, 10427--10434, PMID: 37956397\relax
\mciteBstWouldAddEndPuncttrue
\mciteSetBstMidEndSepPunct{\mcitedefaultmidpunct}
{\mcitedefaultendpunct}{\mcitedefaultseppunct}\relax
\EndOfBibitem
\bibitem[Pak and Farimani(2025)Pak, and Farimani]{pak2025}
Pak,~P.; Farimani,~A.~B. AdditiveLLM: Large Language Models Predict Defects in Additive Manufacturing. 2025; \url{https://arxiv.org/abs/2501.17784}\relax
\mciteBstWouldAddEndPuncttrue
\mciteSetBstMidEndSepPunct{\mcitedefaultmidpunct}
{\mcitedefaultendpunct}{\mcitedefaultseppunct}\relax
\EndOfBibitem
\bibitem[Chaudhari \latin{et~al.}(2024)Chaudhari, Guntuboina, Huang, and Farimani]{chaudhari2024alloybert}
Chaudhari,~A.; Guntuboina,~C.; Huang,~H.; Farimani,~A.~B. AlloyBERT: Alloy property prediction with large language models. \emph{Computational Materials Science} \textbf{2024}, \emph{244}, 113256\relax
\mciteBstWouldAddEndPuncttrue
\mciteSetBstMidEndSepPunct{\mcitedefaultmidpunct}
{\mcitedefaultendpunct}{\mcitedefaultseppunct}\relax
\EndOfBibitem
\bibitem[Jacobs \latin{et~al.}(2024)Jacobs, Polak, Schultz, Mahdavi, Honavar, and Morgan]{jacobs2024regression}
Jacobs,~R.; Polak,~M.~P.; Schultz,~L.~E.; Mahdavi,~H.; Honavar,~V.; Morgan,~D. Regression with Large Language Models for Materials and Molecular Property Prediction. 2024; \url{https://arxiv.org/abs/2409.06080}\relax
\mciteBstWouldAddEndPuncttrue
\mciteSetBstMidEndSepPunct{\mcitedefaultmidpunct}
{\mcitedefaultendpunct}{\mcitedefaultseppunct}\relax
\EndOfBibitem
\bibitem[Chandrasekhar \latin{et~al.}(2024)Chandrasekhar, Chan, Ogoke, Ajenifujah, and {Barati Farimani}]{chandrasekhar2024amgpt}
Chandrasekhar,~A.; Chan,~J.; Ogoke,~F.; Ajenifujah,~O.; {Barati Farimani},~A. AMGPT: A large language model for contextual querying in additive manufacturing. \emph{Additive Manufacturing Letters} \textbf{2024}, \emph{11}, 100232\relax
\mciteBstWouldAddEndPuncttrue
\mciteSetBstMidEndSepPunct{\mcitedefaultmidpunct}
{\mcitedefaultendpunct}{\mcitedefaultseppunct}\relax
\EndOfBibitem
\bibitem[Liu(2019)]{liu2019roberta}
Liu,~Y. Roberta: A robustly optimized bert pretraining approach. \emph{arXiv preprint arXiv:1907.11692} \textbf{2019}, \emph{364}\relax
\mciteBstWouldAddEndPuncttrue
\mciteSetBstMidEndSepPunct{\mcitedefaultmidpunct}
{\mcitedefaultendpunct}{\mcitedefaultseppunct}\relax
\EndOfBibitem
\bibitem[Raffel \latin{et~al.}(2020)Raffel, Shazeer, Roberts, Lee, Narang, Matena, Zhou, Li, and Liu]{raffel2020exploring}
Raffel,~C.; Shazeer,~N.; Roberts,~A.; Lee,~K.; Narang,~S.; Matena,~M.; Zhou,~Y.; Li,~W.; Liu,~P.~J. Exploring the limits of transfer learning with a unified text-to-text transformer. \emph{Journal of Machine Learning Research} \textbf{2020}, \emph{21}, 1--67\relax
\mciteBstWouldAddEndPuncttrue
\mciteSetBstMidEndSepPunct{\mcitedefaultmidpunct}
{\mcitedefaultendpunct}{\mcitedefaultseppunct}\relax
\EndOfBibitem
\bibitem[{Meta AI}(2024)]{meta_llama_32}
{Meta AI} LLaMA 3.2 Model Overview. 2024; \url{https://arxiv.org/pdf/2407.21783}\relax
\mciteBstWouldAddEndPuncttrue
\mciteSetBstMidEndSepPunct{\mcitedefaultmidpunct}
{\mcitedefaultendpunct}{\mcitedefaultseppunct}\relax
\EndOfBibitem
\bibitem[{Meta AI}(2023)]{meta_llama_31}
{Meta AI} LLaMA 3.1 Technical Report. 2023; \url{https://arxiv.org/pdf/2302.13971}\relax
\mciteBstWouldAddEndPuncttrue
\mciteSetBstMidEndSepPunct{\mcitedefaultmidpunct}
{\mcitedefaultendpunct}{\mcitedefaultseppunct}\relax
\EndOfBibitem
\bibitem[Gupta \latin{et~al.}(2022)Gupta, Zaki, Krishnan, and Mausam]{gupta2022matscibert}
Gupta,~T.; Zaki,~M.; Krishnan,~N.~A.; Mausam MatSciBERT: A materials domain language model for text mining and information extraction. \emph{npj Computational Materials} \textbf{2022}, \emph{8}, 102\relax
\mciteBstWouldAddEndPuncttrue
\mciteSetBstMidEndSepPunct{\mcitedefaultmidpunct}
{\mcitedefaultendpunct}{\mcitedefaultseppunct}\relax
\EndOfBibitem
\bibitem[Gossett \latin{et~al.}(2018)Gossett, Toher, Oses, Isayev, Legrain, Rose, Zurek, Carrete, Mingo, Tropsha, \latin{et~al.} others]{gossett2018aflow}
Gossett,~E.; Toher,~C.; Oses,~C.; Isayev,~O.; Legrain,~F.; Rose,~F.; Zurek,~E.; Carrete,~J.; Mingo,~N.; Tropsha,~A.; others AFLOW-ML: A RESTful API for machine-learning predictions of materials properties. \emph{Computational Materials Science} \textbf{2018}, \emph{152}, 134--145\relax
\mciteBstWouldAddEndPuncttrue
\mciteSetBstMidEndSepPunct{\mcitedefaultmidpunct}
{\mcitedefaultendpunct}{\mcitedefaultseppunct}\relax
\EndOfBibitem
\bibitem[Taylor \latin{et~al.}(2014)Taylor, Rose, Toher, Levy, Yang, Nardelli, and Curtarolo]{taylor2014restful}
Taylor,~R.~H.; Rose,~F.; Toher,~C.; Levy,~O.; Yang,~K.; Nardelli,~M.~B.; Curtarolo,~S. A RESTful API for exchanging materials data in the AFLOWLIB. org consortium. \emph{Computational materials science} \textbf{2014}, \emph{93}, 178--192\relax
\mciteBstWouldAddEndPuncttrue
\mciteSetBstMidEndSepPunct{\mcitedefaultmidpunct}
{\mcitedefaultendpunct}{\mcitedefaultseppunct}\relax
\EndOfBibitem
\bibitem[Setyawan \latin{et~al.}(2011)Setyawan, Gaume, Lam, Feigelson, and Curtarolo]{setyawan2011high}
Setyawan,~W.; Gaume,~R.~M.; Lam,~S.; Feigelson,~R.~S.; Curtarolo,~S. High-throughput combinatorial database of electronic band structures for inorganic scintillator materials. \emph{ACS combinatorial science} \textbf{2011}, \emph{13}, 382--390\relax
\mciteBstWouldAddEndPuncttrue
\mciteSetBstMidEndSepPunct{\mcitedefaultmidpunct}
{\mcitedefaultendpunct}{\mcitedefaultseppunct}\relax
\EndOfBibitem
\bibitem[Wang \latin{et~al.}(2022)Wang, Zhang, Th{\'e}, and Yu]{wang2022accurate}
Wang,~T.; Zhang,~K.; Th{\'e},~J.; Yu,~H. Accurate prediction of band gap of materials using stacking machine learning model. \emph{Computational Materials Science} \textbf{2022}, \emph{201}, 110899\relax
\mciteBstWouldAddEndPuncttrue
\mciteSetBstMidEndSepPunct{\mcitedefaultmidpunct}
{\mcitedefaultendpunct}{\mcitedefaultseppunct}\relax
\EndOfBibitem
\bibitem[Tripathy and Pattanaik(2016)Tripathy, and Pattanaik]{tripathy2016optical}
Tripathy,~S.~K.; Pattanaik,~A. Optical and electronic properties of some semiconductors from energy gaps. \emph{Optical Materials} \textbf{2016}, \emph{53}, 123--133\relax
\mciteBstWouldAddEndPuncttrue
\mciteSetBstMidEndSepPunct{\mcitedefaultmidpunct}
{\mcitedefaultendpunct}{\mcitedefaultseppunct}\relax
\EndOfBibitem
\bibitem[He \latin{et~al.}(2018)He, Cubuk, Allendorf, and Reed]{he2018metallic}
He,~Y.; Cubuk,~E.~D.; Allendorf,~M.~D.; Reed,~E.~J. Metallic metal--organic frameworks predicted by the combination of machine learning methods and ab initio calculations. \emph{The journal of physical chemistry letters} \textbf{2018}, \emph{9}, 4562--4569\relax
\mciteBstWouldAddEndPuncttrue
\mciteSetBstMidEndSepPunct{\mcitedefaultmidpunct}
{\mcitedefaultendpunct}{\mcitedefaultseppunct}\relax
\EndOfBibitem
\bibitem[Khan \latin{et~al.}(2023)Khan, Tayara, and Chong]{khan2023prediction}
Khan,~A.; Tayara,~H.; Chong,~K.~T. Prediction of organic material band gaps using graph attention network. \emph{Computational Materials Science} \textbf{2023}, \emph{220}, 112063\relax
\mciteBstWouldAddEndPuncttrue
\mciteSetBstMidEndSepPunct{\mcitedefaultmidpunct}
{\mcitedefaultendpunct}{\mcitedefaultseppunct}\relax
\EndOfBibitem
\bibitem[Wei and Zunger(1998)Wei, and Zunger]{wei1998calculated}
Wei,~S.-H.; Zunger,~A. Calculated natural band offsets of all II--VI and III--V semiconductors: Chemical trends and the role of cation d orbitals. \emph{Applied Physics Letters} \textbf{1998}, \emph{72}, 2011--2013\relax
\mciteBstWouldAddEndPuncttrue
\mciteSetBstMidEndSepPunct{\mcitedefaultmidpunct}
{\mcitedefaultendpunct}{\mcitedefaultseppunct}\relax
\EndOfBibitem
\bibitem[Huang \latin{et~al.}(2019)Huang, Yu, Chen, Liu, Li, Niu, Wang, and Jia]{huang2019band}
Huang,~Y.; Yu,~C.; Chen,~W.; Liu,~Y.; Li,~C.; Niu,~C.; Wang,~F.; Jia,~Y. Band gap and band alignment prediction of nitride-based semiconductors using machine learning. \emph{Journal of Materials Chemistry C} \textbf{2019}, \emph{7}, 3238--3245\relax
\mciteBstWouldAddEndPuncttrue
\mciteSetBstMidEndSepPunct{\mcitedefaultmidpunct}
{\mcitedefaultendpunct}{\mcitedefaultseppunct}\relax
\EndOfBibitem
\bibitem[Hu(2010)]{hu2010modern}
Hu,~C. Modern semiconductor devices for integrated circuits. \emph{(No Title)} \textbf{2010}, \relax
\mciteBstWouldAddEndPunctfalse
\mciteSetBstMidEndSepPunct{\mcitedefaultmidpunct}
{}{\mcitedefaultseppunct}\relax
\EndOfBibitem
\bibitem[Yuan \latin{et~al.}(2018)Yuan, Deng, Li, Wei, and Luo]{yuan2018unified}
Yuan,~L.-D.; Deng,~H.-X.; Li,~S.-S.; Wei,~S.-H.; Luo,~J.-W. Unified theory of direct or indirect band-gap nature of conventional semiconductors. \emph{Physical Review B} \textbf{2018}, \emph{98}, 245203\relax
\mciteBstWouldAddEndPuncttrue
\mciteSetBstMidEndSepPunct{\mcitedefaultmidpunct}
{\mcitedefaultendpunct}{\mcitedefaultseppunct}\relax
\EndOfBibitem
\bibitem[Zheng \latin{et~al.}(2011)Zheng, Cohen, Mori-S{\'a}nchez, Hu, and Yang]{zheng2011improving}
Zheng,~X.; Cohen,~A.~J.; Mori-S{\'a}nchez,~P.; Hu,~X.; Yang,~W. Improving band gap prediction in density functional theory from molecules to solids. \emph{Physical review letters} \textbf{2011}, \emph{107}, 026403\relax
\mciteBstWouldAddEndPuncttrue
\mciteSetBstMidEndSepPunct{\mcitedefaultmidpunct}
{\mcitedefaultendpunct}{\mcitedefaultseppunct}\relax
\EndOfBibitem
\bibitem[Na \latin{et~al.}(2020)Na, Jang, Lee, and Chang]{na2020tuplewise}
Na,~G.~S.; Jang,~S.; Lee,~Y.-L.; Chang,~H. Tuplewise material representation based machine learning for accurate band gap prediction. \emph{The Journal of Physical Chemistry A} \textbf{2020}, \emph{124}, 10616--10623\relax
\mciteBstWouldAddEndPuncttrue
\mciteSetBstMidEndSepPunct{\mcitedefaultmidpunct}
{\mcitedefaultendpunct}{\mcitedefaultseppunct}\relax
\EndOfBibitem
\bibitem[Vasseur \latin{et~al.}(2013)Vasseur, Fagot-Revurat, Kierren, Sicot, and Malterre]{vasseur2013effect}
Vasseur,~G.; Fagot-Revurat,~Y.; Kierren,~B.; Sicot,~M.; Malterre,~D. Effect of symmetry breaking on electronic band structure: gap opening at the high symmetry points. \emph{Symmetry} \textbf{2013}, \emph{5}, 344--354\relax
\mciteBstWouldAddEndPuncttrue
\mciteSetBstMidEndSepPunct{\mcitedefaultmidpunct}
{\mcitedefaultendpunct}{\mcitedefaultseppunct}\relax
\EndOfBibitem
\end{mcitethebibliography}
\end{document}

\tableofcontents
%\clearpage
% \appendix
% \counterwithin{figure}{section}

\newpage
\section{String and Description Examples}

\begin{tcolorbox}[title=\ce{CrO3Ta}]
\textbf{String:}\\
compound: Cr1O3Ta1, species: ['Cr', 'O', 'Ta'], composition: [1, 3, 1], density: 7.274, valence\_cell\_iupac: 17, species\_pp: ['Cr\_pv', 'O', 'Ta\_pv'], spinD: [4.555, 0.016, 0.021, 0.008, 0.061], spin\_atom: 1.0, spin\_cell: 5.001, crystal\_class: hexoctahedral, crystal\_family: cubic, crystal\_system: cubic, positions\_fractional: [[0, 0, 0], [0, 0.5, 0.5], [0.5, 0, 0.5], [0.5, 0.5, 0], [0.5, 0.5, 0.5]], geometry: [4.003, 4.003, 4.003, 90, 90, 90], lattice\_system\_relax: cubic, lattice\_variation\_relax: CUB, spacegroup\_relax: 221, sg: ['Pm-3m \#221', 'Pm-3m \#221', 'Pm-3m \#221'], sg2: ['Pm-3m \#221', 'Pm-3m \#221', 'Pm-3m \#221'], point\_group\_orbifold: *432, point\_group\_order: 48, point\_group\_structure: 2\_x\_symmetric, point\_group\_type: centrosymmetric

\vspace{1em}
\textbf{Description:}\\
The compound CrO$_3$Ta features a composition of 1 Cr atom, 3 O atoms, and 1 Ta atom, with a density of 7.27364 g/cm$^3$. It belongs to the cubic crystal system and has a hexoctahedral crystal class. The lattice system is relaxed cubic with lattice parameters $a = b = c = 4.00287$ Å and angles $\alpha = \beta = \gamma = 90^\circ$. 

The compound has a valence of 17 according to the IUPAC system and crystallizes in the space group Pm-3m \#221. It exhibits spin values of 1.00013 at the atomic level and 5.00063 at the cell level. The structure is highly symmetric, belonging to the *432 point group with an order of 48 and a centrosymmetric configuration. 

The atoms are located at fractional coordinates: (0, 0, 0), (0, 0.5, 0.5), (0.5, 0, 0.5), (0.5, 0.5, 0), and (0.5, 0.5, 0.5). The atomic species present are Cr, O, and Ta, with the pseudopotential designations Cr\_pv, O, and Ta\_pv, respectively.
\end{tcolorbox}

\begin{tcolorbox}[title=\ce{Bi1Dy1Ni1}]
\textbf{String:}\\
compound: Bi1Dy1Ni1, species: ['Bi', 'Dy', 'Ni'], composition: [1, 1, 1], density: 10.599, valence\_cell\_iupac: 13, species\_pp: ['Bi\_d', 'Dy\_3', 'Ni\_pv'], spinD: [0, 0, 0], spin\_atom: 0.0, spin\_cell: 0.0, crystal\_class: tetrahedral, crystal\_family: cubic, crystal\_system: cubic, positions\_fractional: [[0, 0, 0], [0.5, 0.5, 0.5], [0.25, 0.25, 0.25]], geometry: [4.568, 4.568, 4.568, 60, 60, 60], lattice\_system\_relax: cubic, lattice\_variation\_relax: FCC, spacegroup\_relax: 216, sg: ['F-43m \#216', 'F-43m \#216', 'F-43m \#216'], sg2: ['F-43m \#216', 'F-43m \#216', 'F-43m \#216'], point\_group\_orbifold: *332, point\_group\_order: 24, point\_group\_structure: symmetric, point\_group\_type: none

\vspace{1em}
\textbf{Description:}\\
This material is a cubic compound with the chemical formula BiDyNi. It has a density of 10.5987 g/cm$^3$ and a valence of 13 according to the IUPAC system. The crystal structure is tetrahedral within the cubic crystal family and system. The lattice system is relaxed cubic with a face-centered cubic (FCC) lattice variation. The space group is F-43m \#216, and the point group is *332 with an order of 24, showing a symmetric structure. The atomic positions in the unit cell are at (0,0,0), (0.5,0.5,0.5), and (0.25,0.25,0.25). The atomic species present are Bi, Dy, and Ni, with spins of 0 for each atom. The geometry of the unit cell is characterized by lattice parameters of $a=b=c=4.567913$ Å and $\alpha=\beta=\gamma=60^\circ$. The species have the configurations Bi\_d, Dy\_3, and Ni\_pv respectively.
\end{tcolorbox}

\begin{tcolorbox}[title=\ce{Au2Bi2Li4}]
\textbf{String:}\\
compound: Au2Bi2Li4, species: ['Au', 'Bi', 'Li'], composition: [2, 2, 4], density: 8.068, valence\_cell\_iupac: 24, species\_pp: ['Au', 'Bi\_d', 'Li\_sv'], spinD: [0, 0, 0, 0, 0, 0, 0, 0], spin\_atom: 0.0, spin\_cell: 0.0, crystal\_class: orthorhombic-bipyramidal, crystal\_family: orthorhombic, crystal\_system: orthorhombic, positions\_fractional: [[0, 0, 0], [0, 0, 0.5], [0.662, 0.662, 0.25], [0.338, 0.338, 0.75], [0.474, 0.12, 0.25], [0.526, 0.88, 0.75], [0.12, 0.474, 0.25], [0.88, 0.526, 0.75]], geometry: [5.563, 5.563, 5.638, 90, 90, 97.937], lattice\_system\_relax: orthorhombic, lattice\_variation\_relax: ORCC, spacegroup\_relax: 63, sg: ['Cmcm \#63', 'Cmcm \#63', 'Cmcm \#63'], sg2: ['Cmcm \#63', 'Cmcm \#63', 'Cmcm \#63'], point\_group\_orbifold: *222, point\_group\_order: 8, point\_group\_structure: 2\_x\_dihedral, point\_group\_type: centrosymmetric

\vspace{1em}
\textbf{Description:}\\
The compound Au$_2$Bi$_2$Li$_4$ features a unique orthorhombic crystal structure with a crystal class of orthorhombic-bipyramidal, belonging to the orthorhombic crystal family and system. The material has a density of 8.06805 g/cm$^3$ and a valence cell of 24. The chemical composition consists of 2 atoms of Au, 2 atoms of Bi, and 4 atoms of Li. The lattice system is orthorhombic, with lattice parameters $a = b = 5.562936$ Å, $c = 5.638398$ Å, and angles $\alpha = \beta = 90^\circ$, $\gamma = 97.937^\circ$. The space group is Cmcm \#63, with a relaxed lattice system of orthorhombic and lattice variation of ORCC.

The atoms are positioned in the crystal structure at fractional coordinates:
(0, 0, 0), (0, 0, 0.5), (0.662, 0.662, 0.25), (0.338, 0.338, 0.75), (0.474, 0.12, 0.25), (0.526, 0.88, 0.75), (0.12, 0.474, 0.25), and (0.88, 0.526, 0.75).

The point group characteristics include a point group orbifold of *222, an order of 8, a structure of 2\_x\_dihedral, and a centrosymmetric type. Spin properties indicate zero spin for both individual atoms and the overall unit cell.
\end{tcolorbox}

\begin{tcolorbox}[title=\ce{Ag2CrPt}]
\textbf{String:}\\
compound: Ag2Cr1Pt1, species: ['Ag', 'Cr', 'Pt'], composition: [2, 1, 1], density: 12.174, valence\_cell\_iupac: 20, species\_pp: ['Ag', 'Cr\_pv', 'Pt'], spinD: [0, 0, 0, 0], spin\_atom: 0.0, spin\_cell: 0.0, crystal\_class: tetrahedral, crystal\_family: cubic, crystal\_system: cubic, positions\_fractional: [[0, 0, 0], [0.25, 0.25, 0.25], [0.5, 0.5, 0.5], [0.75, 0.75, 0.75]], geometry: [4.469, 4.469, 4.469, 60, 60, 60], lattice\_system\_relax: cubic, lattice\_variation\_relax: FCC, spacegroup\_relax: 216, sg: ['F-43m \#216', 'F-43m \#216', 'F-43m \#216'], sg2: ['F-43m \#216', 'F-43m \#216', 'F-43m \#216'], point\_group\_orbifold: *332, point\_group\_order: 24, point\_group\_structure: symmetric, point\_group\_type: none

\vspace{1em}
\textbf{Description:}\\
The material is a compound with the chemical formula Ag$_2$CrPt, consisting of silver (Ag), chromium (Cr), and platinum (Pt) in a ratio of 2:1:1. It has a density of 12.1739 g/cm$^3$ and a valence of 20 according to the IUPAC standard. The crystal structure is cubic with a tetrahedral arrangement. The lattice system is cubic with a face-centered cubic (FCC) lattice variation, and the space group is F-43m \#216. The point group is *332, indicating a symmetric structure with 24-fold rotational symmetry. 

The system exhibits no net magnetic moment, with spin values of 0.0 at both the atom and cell levels. Atomic positions are defined by fractional coordinates: (0, 0, 0), (0.25, 0.25, 0.25), (0.5, 0.5, 0.5), and (0.75, 0.75, 0.75) within the unit cell. The crystal geometry is characterized by lattice parameters $a = b = c = 4.4694$ Å and angles $\alpha = \beta = \gamma = 60^\circ$.
\end{tcolorbox}

% \bibliography{reference}